\documentclass[conference]{IEEEtran}
\IEEEoverridecommandlockouts
\usepackage{cite}
\usepackage{amsmath,amssymb,amsfonts}
\usepackage{graphicx}
\usepackage{textcomp}
\usepackage{xcolor}
\usepackage{cite}
\def\BibTeX{{\\rm B\\kern-.05em{\\sc i\\kern-.025em b}\\kern-.08em
    T\\kern-.1667em\\lower.7ex\\hbox{E}\\kern-.125emX}}
\begin{document}

\title{Voice-based AI Agents: Filling the Economic Gaps in Digital Health Delivery}

\author{
\IEEEauthorblockN{
Bo Wen\IEEEauthorrefmark{1},
Chen Wang\IEEEauthorrefmark{2},
Qiwei Han\IEEEauthorrefmark{3},
Raquel Norel\IEEEauthorrefmark{1}, \\
Julia Liu\IEEEauthorrefmark{4},
Thaddeus Stappenbeck\IEEEauthorrefmark{5},
Jeffrey L. Rogers\IEEEauthorrefmark{1}
}
\\
\IEEEauthorblockA{
\IEEEauthorrefmark{1}\textit{Digital Health, IBM T.J. Watson Research Center, Yorktown Heights, NY USA}\\
\IEEEauthorrefmark{2}\textit{AI Cloud Platform, IBM T.J. Watson Research Center, Yorktown Heights, NY USA}\\
\IEEEauthorrefmark{3}\textit{Nova School of Business and Economics, Carcavelos, Portugal}\\
\IEEEauthorrefmark{4}\textit{Gastroenterology, Morehouse School of Medicine, Atlanta, GA USA}\\
\IEEEauthorrefmark{5}\textit{Inflammation and Immunity, Cleveland Clinic Foundation, Cleveland, OH USA}\\
\\
Emails: \{bwen, rnorel, jeffrogers\}@us.ibm.com, \\
chen.wang1@ibm.com, qiwei.han@novasbe.pt, \\
jjliu@msm.edu, stappet@ccf.org
}}

\maketitle

\begin{abstract}
    The integration of voice-based AI agents in healthcare presents a transformative opportunity to bridge economic and accessibility gaps in digital health delivery. This paper explores the role of large language model (LLM)-powered voice assistants in enhancing preventive care and continuous patient monitoring, particularly in underserved populations. Drawing insights from the development and pilot study of Agent PULSE (Patient Understanding and Liaison Support Engine)—a collaborative initiative between IBM Research, Cleveland Clinic Foundation, and Morehouse School of Medicine—we present an economic model demonstrating how AI agents can provide cost-effective healthcare services where human intervention is economically unfeasible. Our pilot study with 33 inflammatory bowel disease patients revealed that 70\% expressed acceptance of AI-driven monitoring, with 37\% preferring it over traditional modalities. Technical challenges, including real-time conversational AI processing, integration with healthcare systems, and privacy compliance, are analyzed alongside policy considerations surrounding regulation, bias mitigation, and patient autonomy. Our findings suggest that AI-driven voice agents not only enhance healthcare scalability and efficiency but also improve patient engagement and accessibility. For healthcare executives, our cost-utility analysis demonstrates huge potential savings for routine monitoring tasks, while technologists can leverage our framework to prioritize improvements yielding the highest patient impact. By addressing current limitations and aligning AI development with ethical and regulatory frameworks, voice-based AI agents can serve as a critical entry point for equitable, sustainable digital healthcare solutions.
\end{abstract}


\section{Introduction}
Healthcare systems worldwide face growing challenges in allocating limited medical resources to meet increasing demand~\cite{crowley2020envisioning,khan2024public}. Traditional healthcare delivery models, centered on episodic patient-provider interactions, often result in significant gaps in continuous care, particularly in preventive health monitoring and chronic disease management~\cite{khan2024public,milani2015health}. These shortcomings disproportionately affect vulnerable populations, including those with limited access to healthcare facilities~\cite{keesara2020covid}, lower technological literacy~\cite{nutbeam2021understanding}, or socio-economic constraints~\cite{schillinger2021social}.

The advent of Large Language Models (LLMs) and multimodal AI has opened new avenues for digital health applications~\cite{bedi2024testing,clusmann2023future,lu2024multimodal,thirunavukarasu2023large}, notably in voice-based patient engagement~\cite{mirza2024reclaiming,yang2024talk2care}. Unlike earlier rule-based conversational agents, modern AI-driven voice assistants can facilitate context-aware, adaptive, and natural conversations that dynamically adjust to user preferences, health literacy levels, and immediate needs~\cite{zhan2024healthcare}. Voice, as humanity's most intuitive mode of communication, reduces engagement barriers and broadens access to healthcare, especially for underserved communities~\cite{berube2021voice,yang2024talk2care}. 

For healthcare providers, these technologies promise to extend clinical capabilities beyond facility walls while optimizing resource allocation. A primary care physician managing thousands of patients can realistically engage with only dozens per day, leaving a vast monitoring gap that technology must fill. Meanwhile, technologists seek platforms where AI advances can deliver measurable health outcomes and sustainable business models. At this intersection lies the potential for voice-based AI agents to transform healthcare delivery economics.

This paper presents our viewpoint on how voice-based AI agents can help bridge significant economic gaps in healthcare delivery. We argue that voice interaction, combined with modern AI capabilities, provides an effective ``entry point" for healthcare services, enabling scalable, cost-effective, and equitable solutions. Based on our clinical trial experience with Agent \textbf{PULSE} (\textbf{P}atient \textbf{U}nderstanding and \textbf{L}iaison \textbf{S}upport \textbf{E}ngine)—developed by IBM Research and validated in collaboration with the Cleveland Clinic Foundation and Morehouse School of Medicine—we outline an economic model that demonstrates where AI agents can enhance patient monitoring in scenarios where human medical expertise is either unavailable or economically unjustifiable.

We also highlight the technical challenges, scaling opportunities, and policy implications associated with deploying AI-powered voice agents in digital health ecosystems. For clinicians and healthcare administrators, we provide actionable insights on implementation pathways and economic models. For technologists, we outline technical optimization priorities that directly impact healthcare delivery quality. For policymakers, we identify regulatory considerations that balance innovation with patient safety. This perspective aims to inform research, guide development, and shape policies to create accessible, cost-efficient, and patient-centered AI health solutions.

\section{Voice-based AI: A New Entry Point}

\subsection{The Natural Interface for Human Interaction}
Human beings have evolved to communicate primarily through speech, making voice our most intuitive form of interaction. Despite technological advancements, text and graphical user interfaces often require learned skills and can present barriers for many populations, such as older adults~\cite{pradhan2020use,wilson2021barriers} and people with disabilities~\cite{billah2017ubiquitous,vieira2022impact}. The existence of user experience design as a profession indicates that website and smartphone interfaces are not inherently intuitive for human-AI interaction ~\cite{dudley2018review}. Previous attempts by technology companies to establish voice as a primary interaction channel through smart speakers and voice assistants demonstrated the promise of this approach but fell short due to technological limitations~\cite{ermolina2021voice,kim2021exploring}. Early voice assistants such as Siri and Alexa were limited by rigid command structures and limited understanding of natural language, resulting in frustrating user experiences \cite{ermolina2021voice,natale2021browsing}. 

The emergence of advanced LLMs has fundamentally changed this landscape. Modern LLMs can understand context, maintain coherent conversations over multiple turns, adapt to different communication styles, and generate human-like responses~\cite{yi2024survey}. These capabilities enable truly natural voice interactions that remove many traditional barriers to the use of technology.

\subsection{Replacing Traditional Telephonic Systems}
LLM-driven voice bot technology offers a compelling solution to replace traditional interactive voice response (IVR) systems, which are often plagued by frustrating menu-driven interactions (``press 1 for X, press 2 for Y") that can drive users to bypass the system completely~\cite{shah2024death}. These legacy systems are due to limitations in earlier natural language processing, which could only handle basic speech recognition and simple intent mapping. Engineers were forced to create rigid menu structures that corresponded to specific business logic, resulting in rule-based systems with minimal awareness and personalization of the context. As organizations expanded services, these menu structures became increasingly complex and unintuitive. Users faced multiple layers of options, making voice interfaces less appealing than web or text alternatives. 

LLM-driven systems transform this experience by enabling natural language interactions. Patients can simply state their needs, and AI determines the appropriate service, asks relevant follow-up questions, and provides personalized assistance, eliminating the need for predefined menu navigation~\cite{jo2023understanding}. This approach improves the user experience while improving accessibility for those with limited technological knowledge or physical limitations.

\subsection{Universal Access Through Voice}
The near-universal penetration of telephone technology makes voice-based systems particularly powerful in addressing healthcare disparities~\cite{zhang2019role}. Unlike smartphone apps or web portals that require specific devices, Internet access, and technological proficiency, voice interactions are accessible to virtually anyone with a basic telephone. This universal access is particularly important for elderly populations who may struggle with digital interfaces~\cite{chen2021understanding,pradhan2020use,wilson2021barriers,yang2024talk2care}, people with low literacy or limited English proficiency~\cite{berube2021voice}, people with visual impairments or physical limitations~\cite{billah2017ubiquitous}, low-income communities with limited access to smartphones or computers~\cite{nallam2020question}, and rural populations with unreliable internet connectivity~\cite{sezgin2020scoping}. Using existing telephone infrastructure, voice-based AI agents can reach these populations without requiring additional technology investment or training. 

\section{Filling the Economic Gaps}
\label{sec:economics}
\subsection{Economic Model of Healthcare Resource Allocation}
Healthcare systems worldwide operate within economic constraints where medical resources must be allocated based on severity, necessity, and cost-effectiveness. The substantial investment required to train and maintain qualified medical professionals creates an inherent scarcity, with healthcare systems prioritizing acute care over prevention and continuous monitoring. This creates significant gaps in care delivery that emerge primarily in preventative services and ongoing patient support.

\begin{figure}[htbp]
\centering\includegraphics[width=1\linewidth]{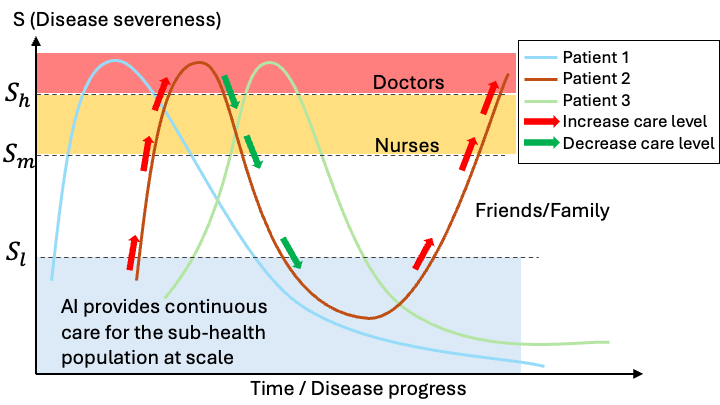}
\caption{Economic model showing the relationship between disease severity/health condition and appropriate care providers across time/disease progression. Colored lines represent individual patients with varying severity trajectories, with green/red arrows indicating transitions between care levels. The blue shaded region illustrates where AI systems can fill gaps in preventative care and monitoring where human medical resources would be economically unjustifiable.}
\label{fig:economics}
\end{figure}

We model this economic reality through a cost-utility analysis framework (Figure \ref{fig:economics}) that illustrates where AI-driven interventions become economically viable. Drawing from Grossman's health capital model \cite{grossman1972concept}, which conceptualizes health as an investment good yielding returns over time, we recognize that preventive interventions generate long-term benefits that are typically undervalued in traditional healthcare financing models.

Let $C_h$ represent the cost of human-provided care (physicians, nurses, clinical staff) and $C_a$ the cost of AI-powered interventions. The cost-efficiency ratio for AI-driven monitoring is given by:
\begin{equation}
    E = \frac{C_h - C_a}{C_h}\times 100\%
\end{equation}

where $E$ represents the percentage reduction in cost achieved by implementing AI-based systems. If $E>0$, AI-based interventions result in cost savings while maintaining service levels.

For a more comprehensive economic evaluation, we can extend this to a cost-utility analysis by incorporating quality-adjusted life years (QALYs) \cite{sinha2022cost}:

\begin{equation}
    \text{ICER} = \frac{C_a - C_h}{\text{QALY}_a - \text{QALY}_h}
\end{equation}

where ICER represents the incremental cost-effectiveness ratio, and QALY$_a$ and QALY$_h$ represent the quality-adjusted life years generated by AI-based and human-based interventions, respectively. This approach allows for a more nuanced understanding of the value provided by AI interventions beyond simple cost reduction.

Figure \ref{fig:economics} illustrates our economic model of healthcare resource allocation across disease progression, inspired by concepts of Pareto efficiency in resource allocation \cite{reinhardt2001economists} and value-based healthcare delivery \cite{porter2010value}. The colored lines represent individual patients (Patient 1, 2, and 3) whose conditions vary in severity over time, with $S$ representing disease severeness. Horizontal thresholds ($S_h$, $S_m$, $S_l$) indicate when different levels of care become economically justifiable:

\begin{itemize}
\item When severity exceeds the high threshold ($S > S_h$), specialized physician care is required
\item For moderate severity ($S_m < S \leq S_h$), nursing care is appropriate
\item For lower severity conditions ($S_l < S \leq S_m$), untrained caregivers or family members typically provide support
\item For mild conditions ($S \leq S_l$), AI monitoring and support is appropriate
\end{itemize}

The figure demonstrates how AI can efficiently fill care gaps, particularly during lower-severity periods when human medical resources would be economically unjustifiable yet monitoring remains beneficial. This model aligns with the concept of risk stratification in healthcare economics \cite{haas2013risk}, where resources are allocated according to patient risk levels and expected benefits. It allows healthcare systems to maximize the utility of scarce physician and nursing resources by focusing them on high-severity cases while deploying AI for continuous monitoring of patients with less severe conditions.

As shown in the trajectories, Patients 1 and 3 maintain relatively stable conditions after their initial treatment, largely due to consistent AI monitoring and support in the mild condition region. In contrast, Patient 2's trajectory shows a pattern of readmission, where the condition deteriorates back to requiring physician care. This illustrates the crucial role of AI in preventive care: by providing continuous monitoring and support during mild conditions, AI helps maintain patient health and prevent hospital readmissions, ultimately reducing the burden on the healthcare system. This demonstrates the phenomenon of ``economies of scope" \cite{panzar1981economies}, where AI systems can efficiently perform multiple monitoring functions at a lower cost than specialized human resources.

This allocation strategy ensures optimal use of scarce medical resources while maintaining quality of care through AI-supported monitoring, particularly in preventing condition deterioration and reducing hospital readmissions.

\subsection{Market Opportunity: AI Filling the Preventative Care Gap}
The absence of scalable preventive care models results in late-stage diagnoses, costly emergency interventions, and avoidable hospitalizations—what health economists term the ``prevention paradox" \cite{rose1981strategy}. AI-driven voice agents offer a solution by providing continuous patient engagement at a fraction of the cost of traditional human-led care.

The key economic advantages of AI-powered healthcare include:

\begin{itemize}
\item \textbf{Economies of scale}: Once deployed, AI voice assistants serve millions of patients with minimal additional marginal cost, exhibiting the classic economic principle where fixed costs are distributed across a larger user base~\cite{haefner2023implementing, besanko2009economics}.
\item \textbf{Continuous availability}: AI systems operate 24/7, eliminating time constraints of human healthcare providers and reducing opportunity costs associated with limited care availability.
\item \textbf{Consistency}: AI removes variability in patient interaction quality, unlike human-driven systems subject to fatigue and workflow constraints, aligning with principles of standardization in service economics \cite{levitt1972production}.
\item \textbf{Proactive Data collection}: AI agents enable early intervention by continuously monitoring symptoms, providing structured patient data for healthcare providers, thereby addressing information asymmetry problems that plague healthcare markets \cite{arrow1963uncertainty}.
\item \textbf{Knowledge updating}: New medical knowledge, practice guidelines, and FDA recommendations can be rapidly integrated through LLM fine-tuning or Retrieval-Augmented Generation (RAG) techniques, making updated information immediately available to all patients without the time and cost of retraining human operators.
\end{itemize}

We estimate the potential cost savings of AI-driven monitoring by evaluating traditional human-driven care models against AI-based interventions. Let $N_p$ be the number of patients requiring monitoring, and let $C_m$ be the cost per patient per month for traditional human-based monitoring. The total cost for human-led monitoring is:
\begin{equation}
C_{human} = N_p \times C_m
\end{equation}
For AI-driven voice agents, the per-patient cost $C_a$ is significantly lower due to automation and scalability, leading to:
\begin{equation}
C_{AI} = F + (N_p \times V_a), \quad \textrm{where} 
\quad V_a \ll C_m
\end{equation}

Where $F$ represents fixed costs of AI system development and deployment, and $V_a$ represents the variable cost per patient. This reflects the economic principle that digital technologies typically have high fixed costs but very low marginal costs \cite{shapiro1999information}, creating a distinctly different economic profile compared to human-delivered services.

The cost reduction factor is thus:
\begin{equation}
    R = \frac{C_{human}-C_{AI}}{C_{human}} \times 100\%
\end{equation}

In a comprehensive cost-benefit analysis framework \cite{drummond2015methods}, the long-term benefits of preventive care include:
\begin{equation}
    NPV = \sum_{t=0}^{T} \frac{B_t - C_t}{(1+r)^t}
\end{equation}

Where $NPV$ is the net present value, $B_t$ represents benefits at time $t$ (including avoided hospitalizations, emergency visits, and productivity gains), $C_t$ represents costs at time $t$, $r$ is the discount rate, and $T$ is the time horizon. This approach recognizes that the benefits of preventive interventions often accrue over longer time periods, while costs are incurred upfront.

By filling this gap, AI-driven patient monitoring can significantly reduce per-patient costs, making preventive care economically viable for larger populations and potentially resolving what health policy experts call the ``value gap" in healthcare delivery \cite{berwick2008triple}.

\subsection{Stakeholder Benefits and Business Models}
The economic gap in preventive care represents a significant market opportunity with distinct benefits for multiple stakeholders:

\textbf{For health insurers}, AI-powered preventive monitoring aligns perfectly with their financial incentives to reduce claims expenses. By minimizing hospitalizations, emergency visits, and complications of chronic diseases, insurers can significantly lower costs while improving outcomes. Voice-based AI health agents offer an efficient population health management tool that proactively engages members before they develop severe conditions.

\textbf{For healthcare providers} shifting towards value-based care models, AI-driven monitoring represents an attractive cost-containment strategy. Under value-based reimbursement, where compensation ties to patient outcomes rather than service volume, AI systems help improve compliance, enable early interventions, and free up clinical resources for higher-priority cases.

\textbf{For healthcare professionals} across various roles, AI agents offer both opportunities and challenges that must be carefully managed:

\begin{itemize}
\item \textbf{Physicians}: AI systems can handle routine monitoring tasks, freeing physicians to focus on complex cases requiring their expertise. However, some may resist adoption due to concerns about patient relationships, liability, or displacement. The key to acceptance is demonstrating how AI enhances rather than replaces their clinical judgment, providing them with better data for decision-making while handling tasks that would otherwise be economically unfeasible.

\item \textbf{Nurses}: While AI can reduce administrative burden and call fatigue (as observed in our MSM case study), nurses may worry about role redefinition. Successful implementation positions AI as handling routine check-ins while elevating nursing work toward more meaningful patient interactions and care coordination that leverages their unique skills.

\item \textbf{Administrative staff}: These professionals benefit from reduced documentation burden and more systematic patient data, though they may require training and workflow adjustments. The economic benefits become most apparent when AI integration reduces overtime and improves staff retention by eliminating frustrating repetitive tasks.

\item \textbf{Support services}: Healthcare ecosystem professionals like IT specialists gain new roles in AI system maintenance, while others may need reassurance about job security. Creating clear upgrade paths for affected roles is essential for organization-wide acceptance.
\end{itemize}

\textbf{For patients}, AI-powered voice assistants democratize healthcare access, transforming personalized health guidance from an exclusive service into a widely available solution. These systems improve accessibility for underserved populations, offer greater privacy and convenience than traditional check-ins, and provide 24/7 access without scheduling constraints.

For successful implementation, healthcare organizations must address professional resistance by demonstrating concrete benefits for each stakeholder group, providing appropriate training, and involving clinical staff in system design and governance. When AI systems are positioned as extending professional capabilities rather than replacing them, they become a "not-not" (inevitable necessity) even for stakeholders who might initially resist adoption \cite{chanoff2023heart}.

These aligned stakeholder benefits create a sustainable business model for AI-driven preventative care that addresses both economic constraints and care quality objectives while respecting the essential roles of all healthcare professionals.

\subsection{From Economic Theory to Practical Implementation}
The economic model presented above provides a compelling theoretical framework for understanding where AI-driven healthcare interventions can fill critical gaps in service delivery. To validate this model and explore its practical implications, we developed Agent PULSE — a voice-based AI system designed to conduct medical surveys and monitor patient conditions through natural conversation.

Agent PULSE serves as a real-world implementation of our economic theory, targeting precisely the ``blue zone" identified in Figure \ref{fig:economics} where human medical resources would be economically unjustifiable yet monitoring remains beneficial. By focusing on routine check-ins and standardized health assessments, this system aims to extend the reach of healthcare services while optimizing resource allocation. The following section details our experience designing, implementing, and testing this system with actual patients, providing empirical insights that complement our theoretical economic framework.

\section{Agent PULSE: Experimental Insights}

\subsection{System Design and Implementation}
Agent PULSE is a telephonic AI system designed to conduct medical surveys and monitor patient conditions through natural conversation. The core of the system is a dialogue management engine powered by prompt-tuned LLMs capable of understanding patient responses, asking follow-up questions, providing guidance and education, and escalating to human healthcare personnel when necessary. A sample conversation demonstrating the system's capabilities is provided in Appendix~\ref{appendix:sample_conversation}.

\begin{figure}[htbp]
\centering\includegraphics[width=1\linewidth]{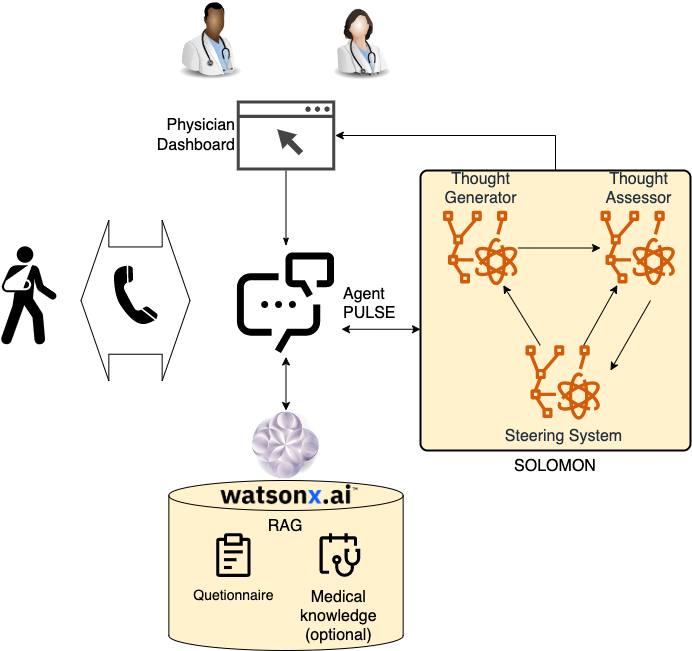}
\caption{Agent PULSE Architecture: The system integrates a voice interface for patient interaction with an advanced AI engine (IBM Watsonx.ai) that leverages both questionnaire data and optional medical knowledge through RAG. The system employs IBM's SOLOMON framework for intelligent conversation management and analysis.}
\label{fig:pulse_architecture}
\end{figure}

As shown in Figure \ref{fig:pulse_architecture}, Agent PULSE's architecture consists of several key components working in concert to deliver intelligent healthcare interactions:

\textbf{Voice Interface}: The system is accessible through standard telephone lines, ensuring broad usability across different demographics, particularly beneficial for patients who may face literacy, technological, or financial barriers to smartphone or computer-based interfaces. Thanks to LLM integration, the voice interface supports multiple languages, providing access to diverse patient populations regardless of their native language.

\textbf{AI Engine}: The core intelligence is powered by IBM's watsonx platform, which integrates:
    \begin{itemize}
    \item \textit{Large Language Models (LLMs)}: Watsonx.ai is a LLM inference service that supports multiple LLM providers, allowing for custom LLM configurations and fine-tuning.
    \item \textit{SOLOMON}: A proprietary IBM multi-LLM-agent reasoning framework that enables intelligent conversation management and automated analysis of unstructured patient conversations~\cite{chen2024comparative, wen2024enhancing}. SOLOMON automatically extracts structured information from natural conversations, converting free-form dialogue into standardized questionnaire responses that care teams can readily review.
    \item \textit{Retrieval-Augmented Generation (RAG)}: Combines structured questionnaire data with optional medical knowledge to inform responses
    \end{itemize}

\textbf{Physician Dashboard}: A comprehensive interface where healthcare providers can:
    \begin{itemize}
    \item Manage calling schedules for large patient cohorts, optimizing resource allocation and ensuring timely follow-ups
    \item Customize survey parameters including language preferences, questionnaire content, time zones, and other logistics to accommodate individual patient needs
    \item Review call results and survey summaries generated by SOLOMON, and plan care interventions accordingly
    \item Track longitudinal patient progress through automated trend analysis of repeated assessments
    \end{itemize}

\subsection{Pilot Study on Patient Engagement}
Prior to implementing Agent PULSE, Morehouse School of Medicine (MSM) faced significant challenges in monitoring patients' conditions between clinical visits. Initially, MSM employed two nurses to conduct individual follow-up calls with patients. However, this approach proved costly and unsustainable, as the nurses quickly experienced burnout due to the high volume of calls. To improve scalability, MSM transitioned to a group-based model using Zoom meetings, where multiple patients simultaneously shared their disease progress with the care team. While this approach allowed providers to reach more patients, it required individuals to discuss personal symptoms and health concerns in the presence of other patients, raising privacy concerns and potentially limiting disclosure of sensitive information.

These challenges highlight the economic gap in healthcare delivery discussed earlier in this paper—where human-led monitoring becomes unsustainable at scale. The cultural and linguistic barriers faced by human providers further exacerbate these inefficiencies, as healthcare systems cannot economically justify hiring culturally matched staff for every patient subgroup. Voice-based AI agents like PULSE represent a potential solution by offering continuous, cost-effective patient monitoring while reducing the burden on healthcare providers and transcending cultural barriers through consistent, non-judgmental interactions.

It is worth noting that our study design deliberately excluded a comparison with traditional menu-driven Interactive Voice Response (IVR) systems (e.g., "press 1 for yes, press 2 for no"). Although such a comparison might have been methodologically interesting from a pure research perspective, our healthcare partners firmly opposed including these legacy systems in the trial. Their resistance stemmed from overwhelming evidence of patient frustration with traditional IVR systems in healthcare settings and concern about recruiting challenges if patients were asked to interact with systems widely considered obsolete. This resistance itself represents significant evidence of the market failure of pre-LLM voice systems in healthcare. Unlike incremental improvements to existing technology, LLM-driven voice agents represent a fundamental paradigm shift that creates entirely new possibilities for patient engagement—addressing needs that previous technologies fundamentally failed to meet.

To evaluate this potential, we conducted a pilot study with a cohort of 33 patients from MSM with inflammatory bowel disease (IBD) who had previously participated in Zoom-based group sessions for condition monitoring. During the trial, patients interacted with Agent PULSE via telephone and were asked the same health assessment questions that were typically covered in their Zoom group sessions. This parallel approach allowed for direct comparison between the two modalities.

\begin{figure}[htbp]
\centering\includegraphics[width=0.8\linewidth]{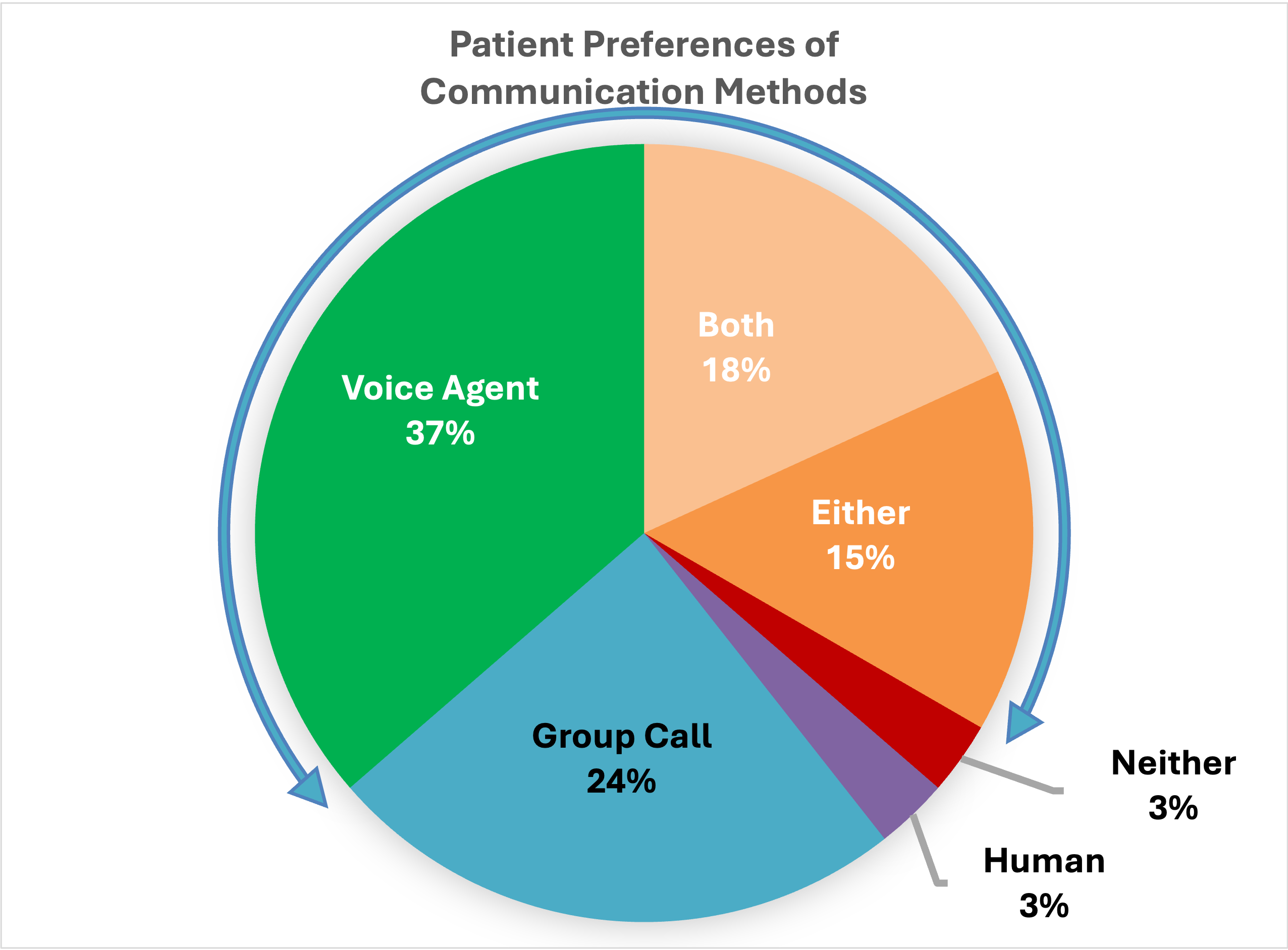}
\caption{Distribution of patient preferences regarding interaction methods. The majority (37\%+18\%+15\%=70\%) expressed acceptance of the AI chatbot communication modality.}
\label{fig:preferences}
\end{figure}

Our findings revealed encouraging patient receptivity to AI-driven healthcare interactions. As shown in Figure \ref{fig:preferences}, 37\% of patients preferred the AI chatbot over other methods, while 24\% favored the Zoom group approach primarily for its social interaction component. Notably, 18\% of patients recognized value in both approaches, suggesting they serve complementary purposes, while 15\% expressed no strong preference between the methods. Only 3\% specifically preferred human interaction, and another 3\% disliked both approaches.

The preference for Agent PULSE among a significant portion of patients may be attributed to several factors:

\begin{itemize}
\item \textbf{Convenience}: Patients could engage with the system at times convenient for them, without scheduling constraints.
\item \textbf{Privacy}: The perception of greater anonymity when discussing sensitive health issues with an AI system versus group settings.
\item \textbf{Reduced judgment}: Patients reported feeling less judged or rushed when communicating with the AI system.
\item \textbf{Consistency}: The AI system provided consistent interaction quality regardless of time of day or frequency of contact.
\end{itemize}

\begin{figure}[htbp]
\centering\includegraphics[width=1\linewidth]{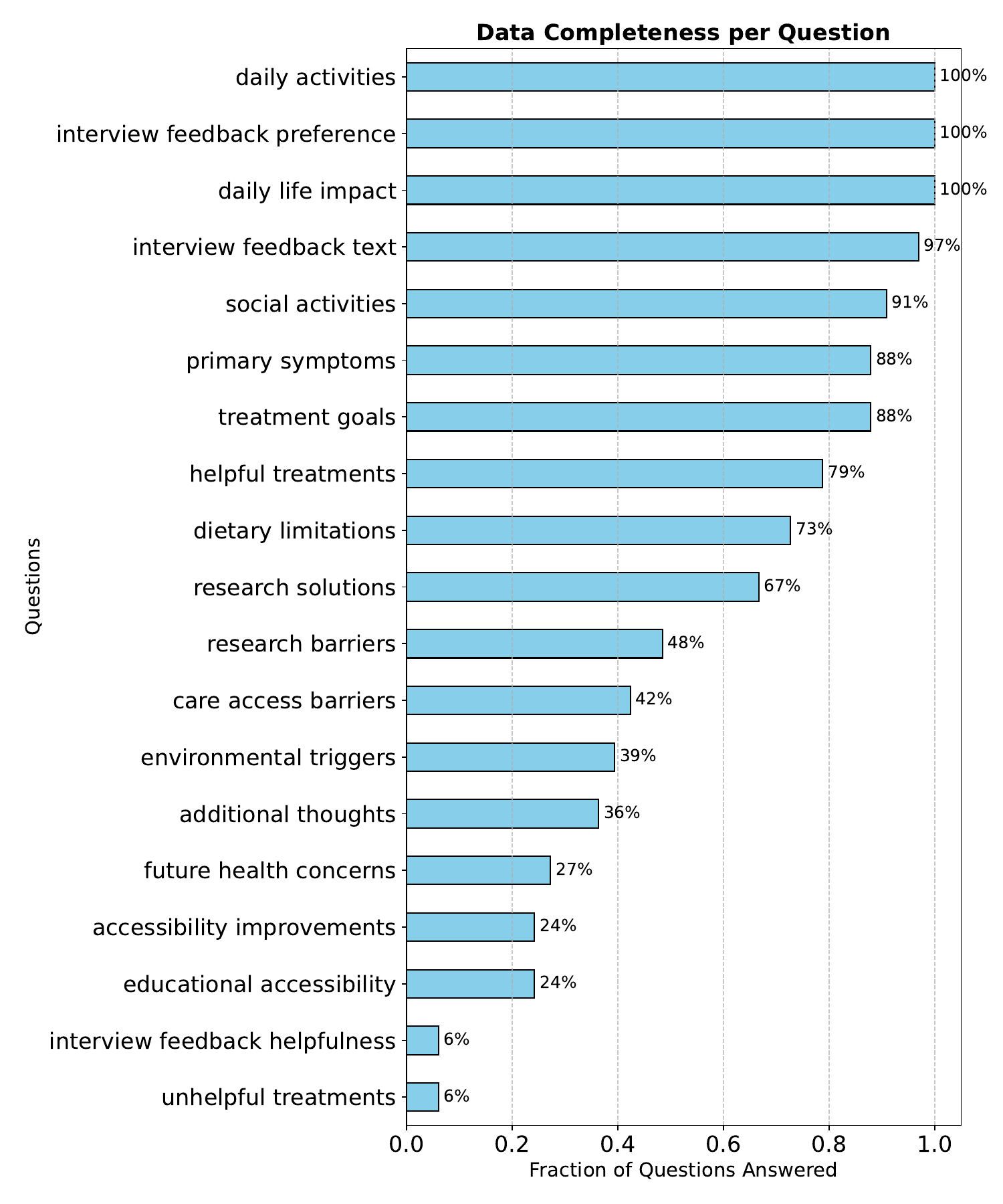}
\caption{Data completeness across different question categories. Questions about daily activities and impact show the highest completion rates ($>$90\%), while more sensitive topics like feedback on treatment helpfulness and interview preferences show significantly lower response rates ($<$10\%).}
\label{fig:completness}
\end{figure}

Our analysis of response completeness (Figure \ref{fig:completness}) revealed significant variations across question categories. Questions about daily activities and daily life impact achieved the highest completion rates (94.4\%), likely because these topics directly relate to patients' symptoms and well-being—areas where patients have strong motivation to provide comprehensive information.

In contrast, questions about research solutions, environmental triggers, and treatment feedback showed substantially lower completion rates, particularly those appearing later in the survey. Through examination of conversation transcripts, we identified several factors contributing to this pattern:

\begin{itemize}
\item \textbf{Question relevance}: Patients were more engaged with questions directly related to their symptoms and treatment than with abstract research-oriented questions that didn't appear to immediately benefit their care.
\item \textbf{Survey fatigue}: Questions positioned later in the survey generally received shorter responses as patient engagement decreased over time.
\item \textbf{Environmental distractions}: In several instances, patients became distracted during calls—in one case by a pet, in another while driving through a toll booth—indicating less commitment to completing the interaction than might be observed with a human provider.
\item \textbf{Perception of AI}: Patients clearly recognized they were interacting with an automated system rather than a human healthcare provider, which influenced their behavior and level of engagement.
\end{itemize}

These observations suggest a significant finding: patients interacted with the AI system differently than they would with human providers. The absence of social pressure or ``white-coat effect" may have allowed patients to express themselves more authentically, potentially revealing more accurate information about their conditions. This has also been observed in \cite{agurto2020detection}: humans behave differently depending on whether they are being observed or not. However, this same dynamic also resulted in less consistent completion of the full assessment. This trade-off between authenticity and completeness represents an important consideration for the design of AI healthcare interactions and warrants further investigation in future studies.

From the healthcare providers' perspective, Agent PULSE offered significant workflow advantages. The system demonstrably reduced administrative burden by automating routine check-ins and standardizing data collection across patients. This standardization was particularly valuable for clinicians, enabling more efficient review of patient status and earlier identification of deteriorating conditions than would be possible with traditional episodic care.

The SOLOMON framework's automated analysis capabilities provided substantial time savings for the healthcare team. By automatically extracting key health metrics from unstructured conversations based on the standard questionnaire, the system enabled quick assessment by providers without requiring them to review lengthy transcripts. Additionally, the built-in emergency detection generated automatic alerts when concerning patterns emerged in patient responses, ensuring critical situations didn't go unnoticed.

The data completeness patterns observed in Figure \ref{fig:completness} also provided valuable insights for care teams. The high response rates for questions about daily activities, symptoms, and treatment goals indicated areas where patients felt comfortable communicating with AI systems. The lower response rates for questions about environmental triggers, care barriers, and accessibility concerns highlighted areas where patients might benefit from human provider follow-up, as well as future survey design considerations.

The pilot study also identified several opportunities for improvement. From a user experience perspective, enhancements to the physician dashboard, better visualization of survey results, and more intuitive organization of patient information would improve clinical utility. More fundamental technological challenges included reducing conversation latency and enhancing the LLM's reasoning capabilities to better handle complex or unexpected patient responses. Additionally, better integration with electronic health records, more sophisticated symptom trend analysis over time, and enhanced mechanisms for AI-to-provider handoff when complex clinical judgment is required would further improve the system's value in clinical settings.

Future iterations of Agent PULSE could benefit from incorporating validated survey instruments from other health domains. For instance, validated nutrition assessment tools~\cite{epi_diet_surveys} could enhance the system's ability to collect comprehensive lifestyle data, addressing a critical gap in chronic disease management where dietary information is often too burdensome to collect through traditional methods.

These findings suggest that voice-based AI agents may not only serve as cost-effective alternatives to human providers in certain contexts but may also be preferred by patients in some scenarios due to their convenience, privacy, and consistent interaction quality. The efficiency gains observed for healthcare providers further support the economic case for AI-driven patient monitoring presented earlier in this paper.

\subsection{Bridging Theory and Practice}
Our experience with Agent PULSE provides empirical support for the economic model presented in Section \ref{sec:economics}. The pilot study clearly demonstrated that voice-based AI agents can effectively fill gaps in care delivery, particularly for routine monitoring between clinical visits. The high acceptance rate among patients (70\% expressing comfort with AI interaction) suggests that such systems can achieve the necessary user engagement to deliver economic benefits in real-world settings.

The challenges faced by Morehouse School of Medicine before implementing Agent PULSE—nurse burnout from high call volumes and privacy concerns with group sessions—directly illustrate the economic inefficiencies our model identifies. Neither approach (individual calls by nurses or group Zoom sessions) offered a sustainable solution for continuous patient monitoring at scale. Agent PULSE addressed this gap by providing a scalable, cost-effective alternative that maintained privacy while reducing provider burden.

These findings validate key aspects of our economic model, particularly the premise that AI systems can efficiently serve patients with lower-severity conditions ($S \leq S_l$ in Figure \ref{fig:economics}), freeing human resources for higher-acuity cases. The technical challenges identified in the following section represent the next critical steps in refining these systems to fully realize the economic potential outlined in our model.

\section{Technical Challenges and Roadmap}

The development and deployment of voice-based AI health agents present several technical challenges that must be addressed to achieve scalable, reliable, and effective systems. 
Based on our experience with Agent PULSE, we identify the following key challenges and opportunities for future research and development.

\subsection{Efficient Conversation Management}
One of the most important factors affecting patient experience with voice AI systems is response time and consistency. When AI systems take too long to respond or fail to remember previous parts of a conversation, patients may become frustrated and disengage. 

Our work identified significant opportunities to improve how AI systems manage conversations through better memory management techniques. These improvements could reduce AI response times by 2-3 times while maintaining natural flow in conversations. For patients, this means fewer awkward pauses and a more natural interaction similar to speaking with a human healthcare provider. For healthcare systems, these optimizations translate to lower operational costs and the ability to serve more patients with existing computing resources.

Technical readers interested in the specific approaches to conversation optimization through Key-Value cache management and other advanced techniques can find detailed explanations in Appendix~\ref{appendix:technical}.

\subsection{Infrastructure Optimization}
Voice-based AI health systems must address two distinct service patterns with different infrastructure requirements:

\begin{itemize}
\item \textbf{Inbound calls}: Similar to web services, inbound calls can experience unpredictable demand spikes, requiring rapid scaling of compute resources.
\item \textbf{Outbound calls}: These can be scheduled during periods of low resource utilization, similar to batch computing tasks.
\end{itemize}

Optimizing infrastructure across these service patterns to maximize resource utilization represents an important area of research in cloud computing. Techniques for mixing different types of services and dynamically allocating resources based on demand patterns could significantly improve the economic viability of these systems~\cite{zhang2024application}.

\subsection{Privacy, Security, and Compliance}
Healthcare applications must comply with regulatory frameworks such as the Health Insurance Portability and Accountability Act (HIPAA) in the United States and the General Data Protection Regulation (GDPR) in Europe, adding complexity to the design of cloud computing architectures~\cite{terry2017existential}. These regulations impose strict requirements for data encryption, access controls, audit trails, and data retention policies~\cite{barati2021privacy,chen2024generative}.

Future research should focus on developing reference architectures and best practices for the deployment of voice-based AI health systems that meet these regulatory requirements while maintaining performance and cost-effectiveness. Techniques such as federated learning, differential privacy, and homomorphic encryption offer promising approaches to balance privacy protection with AI system performance~\cite{piras2019defend,yeng2019framework}.

\subsection{Personalization and Adaptation}
To maximize effectiveness, voice-based AI health agents must adapt to individual patient communication styles, preferences, health literacy levels, and cultural contexts~\cite{ahmad2022designing}. Current methods predominantly employ static prompts that don't fully exploit LLM adaptive capabilities~\cite{abbasian2023conversational}.

A comprehensive personalization framework should include: dynamic patient profiles that evolve based on interaction history; language complexity adjustment matching health literacy levels; cultural competence in conversational strategies; and personalized timing and frequency of outreach based on patient preferences and response patterns. These capabilities would significantly improve engagement, trust, and sustained patient participation while ensuring comprehension and adherence across diverse populations.

\subsection{Retrieval-Augmented Generation}
AI health agents must effectively access and incorporate current medical knowledge and patient-specific data during interactions to deliver accurate, reliable, and evidence-based health guidance. Retrieval-Augmented Generation (RAG) represents a powerful method that combines the generative capabilities of LLM with the structured retrieval of information from specialized domain-specific knowledge bases \cite{ng2025rag}. By integrating retrieval techniques directly into the conversational workflow, AI health agents can generate more informed and precise responses, ensuring that patient guidance aligns closely with established medical guidelines, evidence-based practices, and the most recent clinical research.

Future research efforts should specifically focus on the refinement and enhancement of explicitly designed RAG architectures for healthcare settings. This includes developing sophisticated mechanisms for real-time information retrieval, continuous updates to knowledge bases, and rigorous validation of retrieved content. Ensuring that the medical information recovered remains current, accurate, and clinically relevant is critical, given the dynamic nature of medical knowledge and ongoing updates to clinical guidelines and recommendations.

Furthermore, effective implementation of RAG techniques within AI agents in healthcare will require rigorous mechanisms to verify the credibility and relevance of the retrieved information sources\cite{ng2025rag}. This includes developing quality assurance and validation methods, which may involve the integration of expert-reviewed sources, evidence-grading systems, and transparent documentation of source credibility within conversational responses. Such measures will improve the trustworthiness of AI-generated healthcare information, fostering greater patient confidence and adherence~\cite{macia2024conversational,zhan2024healthcare}.

\section{Conclusion}

Voice-based AI agents powered by LLMs represent a transformative force in healthcare delivery, addressing fundamental economic inefficiencies while democratizing access to preventive care and continuous monitoring. Our research, including clinical insights from the Agent PULSE initiative with 33 IBD patients, demonstrates that these AI-driven systems can fill critical gaps, particularly for populations facing technological and financial barriers to traditional services.

\subsection*{For Healthcare Executives and Administrators}
Our findings present a compelling value proposition: by implementing voice-based AI, organizations can significantly extend care reach while potentially reducing per-patient monitoring costs. The over 70\% patient acceptance rate observed in our study suggests these solutions can achieve the engagement necessary for clinical effectiveness. Strategic implementation should focus initially on chronic condition monitoring and preventive care reminders—areas where our economic model shows the highest return on investment.

\subsection*{For Healthcare Professionals}
Physicians, nurses, and other clinical staff may benefit from understanding how AI voice systems can complement rather than replace their expertise. These systems excel at routine monitoring tasks that would otherwise consume valuable clinical time or remain undone due to resource constraints. By automating these interactions, AI voice systems free healthcare professionals to focus on cases requiring their unique expertise and judgment. When viewed as extending professional capabilities rather than replacing them, AI voice systems become valuable allies in providing comprehensive care across the entire patient population.

\subsection*{For Patients}
Our research demonstrates significant benefits for patients using voice-based AI health systems. These technologies offer unprecedented convenience—allowing health check-ins at any time without scheduling constraints or travel requirements. Patients in our study particularly valued the privacy and reduced judgment when discussing sensitive health issues with AI systems compared to group settings or even human providers. The consistent quality of interaction, regardless of time of day, ensures reliable support when needed most. For underserved populations, including elderly patients, those with limited technological literacy, or individuals in rural areas, voice-based AI removes traditional barriers to healthcare access. By simply using a telephone, patients can receive personalized health monitoring and guidance that would otherwise be unavailable or unaffordable. This democratization of healthcare access represents a fundamental shift toward more equitable health services.

\subsection*{For Technologists and Developers}
Our research identifies priority areas for innovation: (1) optimizing session management and KV cache utilization to improve computing efficiency; (2) developing infrastructure that efficiently handles both scheduled and on-demand patient interactions; and (3) creating personalization frameworks that adapt to individual health literacy, cultural context, and communication preferences. Additionally, integration with existing clinical workflows through standardized APIs represents a critical development priority. For detailed discussion of these technical challenges and their implications for healthcare delivery, see Appendix~\ref{appendix:technical}.

\subsection*{For Policymakers and Regulators}
Beyond technical advances, broader ethical considerations like ensuring AI complements human connection, seamless escalation to human providers, and vigilant monitoring for algorithmic bias, especially given historical disparities, are critical. Voice-based AI offers an inclusive path to democratizing healthcare, transforming care delivery by extending provider reach to underserved populations and addressing economic constraints. Future research should target longitudinal outcomes, remote monitoring integration, and condition-specific applications.

Success requires multidisciplinary collaboration: clinicians contributing expertise, technologists developing robust systems, policymakers establishing supportive frameworks, and patients providing essential feedback. This collaborative approach will maximize voice-based AI's potential as an entry point for equitable, sustainable healthcare delivery.

\section*{Acknowledgment}

We would like to express our sincere gratitude to Jiawen Li, Emem Ekpo, Kennedy Watson and Brian Xu from MSM for their invaluable contributions to this research. Their dedication in managing the patient cohort during the pilot study and providing thoughtful feedback on Agent PULSE significantly enhanced the quality and impact of our work. Their clinical insights and patient management expertise were instrumental in refining the system and ensuring its relevance to real-world healthcare settings. We also extend our appreciation to Craig Caruso, who has recovered from his fight with Crohn's disease, for providing his perspectives as a patient to help us improve the design of the user experience of the chatbot. His firsthand insights were invaluable in creating a more empathetic and effective system. We also acknowledge the use of Claude 3.7 Sonnet, an AI language model by Anthropic, for assistance with grammar, LaTeX formatting, and general English writing improvements throughout the manuscript.

\bibliographystyle{IEEEtran}  
\bibliography{references}

\appendix
\subsection{Sample Conversation and Analysis}\label{appendix:sample_conversation}

Below is an example conversation between Agent PULSE and a patient with inflammatory bowel disease. This conversation demonstrates how the system conducts a health assessment through natural dialogue, gathering information about symptoms and health status in a conversational manner.

This conversation integrates two standard health assessment tools—the Modified Health and Behavioral Index (MHBI) and the EQ-5D-3L health status questionnaire—into a single, streamlined interaction. Rather than conducting separate assessments, Agent PULSE consolidates common question areas while ensuring comprehensive coverage of both instruments. The SOLOMON framework then analyzes the conversation to extract relevant information for both scales simultaneously.

\subsubsection{Conversation Excerpt}
\begin{quote}
\textbf{Assistant:} Hi Craig, this is Agent PULSE from Cleveland Clinic. How are you feeling today, can you rate your overall health from 0 to 100, with 0 being the worst and 100 being the best?

\textbf{Patient:} 25\%.

\textbf{Assistant:} So that's a 25 out of 100. I'm so sorry to hear that you're not feeling your best right now. Can you tell me a little bit about your abdominal pain - are you experiencing any pain in your stomach area today, and if so, how bad is it?

\textbf{Patient:} I'm experiencing some loading. And having some gas issues.

\textbf{Assistant:} It sounds like you're having some discomfort with bloating and gas. Have you had any bowel movements in the past 24 hours, and if so, how many and what was the consistency like?

\textbf{Patient:} Three times. And it's all been diarrhea.

\textbf{Assistant:} I'm so sorry to hear that you're dealing with diarrhea. Have you noticed any other symptoms outside of your stomach, such as joint pain, eye inflammation, or any skin issues like ulcers or bumps?

\textbf{Patient:} Stress and anxiety.

\textbf{Assistant:} It can be really tough to deal with stress and anxiety on top of your physical symptoms. Are you experiencing any discomfort or pain in your private areas, or have you noticed any unusual internal conditions, like rectal pain or discharge?

\textbf{Patient:} No.

\textbf{Assistant:} That's good to know. How about your mobility and daily activities - are you able to get around and do the things you need to do, like taking a shower, getting dressed, and moving around your home, without too much trouble?

\textbf{Patient:} Having joint pain when moving.

\textbf{Assistant:} The joint pain is making it harder for you to move around and do your daily activities. Are you able to take care of yourself, like washing and dressing, without needing too much help from others?

\textbf{Patient:} My wife helps me.

\textbf{Assistant:} It's great that you have your wife's support. How are you feeling emotionally, aside from the stress and anxiety you mentioned earlier - are you feeling overwhelmed, depressed, or hopeless, or are you able to cope with your emotions pretty well?

\textbf{Patient:} Yeah. I was feeling very tired.

\textbf{Assistant:} It sounds like the fatigue is really getting to you. Is there anything else that's been on your mind or any other concerns you'd like to share with me before we wrap up our conversation today?

\textbf{Patient:} First contact to call me back.

\textbf{Assistant:} I'll make sure to have someone from our team follow up with you. We'll schedule a callback for you. Before I let you go, I just want to thank you for taking the time to speak with me today, Craig. I appreciate your honesty and sharing your concerns with me. Is there a specific time that works best for you for the callback, or would you like me to arrange something for tomorrow?

...
\end{quote}

One notable strength of the LLM-powered Agent PULSE is its ability to handle imperfect speech-to-text transcriptions. Even when transcriptions contain errors or inaccuracies (as seen in the patient's response about "loading" instead of "bloating"), the system can still understand the intended meaning and maintain conversational flow. This capability creates a smoother user experience compared to rule-based systems that might fail when encountering unexpected inputs. When patient responses are particularly confusing due to transcription errors, the system intelligently requests clarification rather than proceeding with potentially incorrect assumptions, balancing conversation efficiency with accuracy.

\subsubsection{Automated Analysis Output}
The SOLOMON framework automatically analyzed this conversation and generated the following structured assessment scores:

\begin{table}[htbp]
\caption{Modified Health and Behavioral Index (MHBI) Assessment}
\centering
\begin{tabular}{|p{0.2\linewidth}|p{0.2\linewidth}|p{0.45\linewidth}|}
\hline
\textbf{Measure} & \textbf{Score} & \textbf{Notes} \\
\hline
Liquid Stools & 3 & Patient reported having diarrhea three times in the past 24 hours \\
\hline
Abdominal Pain & 2 & Patient reported bloating and gas issues \\
\hline
General \newline Wellbeing & 3 & Patient rated overall health at 25/100 and reported significant fatigue \\
\hline
Additional \newline Manifestations & 2 & Patient reported stress, anxiety, and joint pain \\
\hline
\end{tabular}
\end{table}

\begin{table}[htbp]
\caption{EQ-5D-3L Health Assessment}
\centering
\begin{tabular}{|p{0.2\linewidth}|p{0.2\linewidth}|p{0.45\linewidth}|}
\hline
\textbf{Dimension} & \textbf{Score} & \textbf{Notes} \\
\hline
Mobility & 2 & Patient reported joint pain when moving \\
\hline
Self-Care & 2 & Patient requires help from spouse for self-care activities \\
\hline
Pain / \newline Discomfort & 2 & Patient reported abdominal discomfort and joint pain \\
\hline
Usual \newline Activities & 2 & Patient has difficulty with daily activities due to joint pain \\
\hline
Anxiety / \newline Depression & 2 & Patient reported stress, anxiety, and fatigue \\
\hline
Health Scale & 25/100 & Patient explicitly rated their health at 25\% \\
\hline
\end{tabular}
\end{table}

\begin{quote}
\textbf{Recommended Action:} Patient requested a callback and follow-up from a human healthcare provider.
\end{quote}

This example demonstrates how Agent PULSE can maintain natural conversations while automatically extracting and structuring critical health information for healthcare providers.

\subsection{Non-Technical Summary of Technical Challenges}\label{appendix:nontechnical}

\subsubsection{Non-Technical Summary: Understanding the Technical Foundations of Voice AI in Healthcare}

For healthcare professionals, it's important to understand what makes voice-based healthcare AI systems work effectively and how potential improvements in these technical areas could directly benefit patients and providers in the future.

\textbf{The Basics:} Voice AI healthcare systems must perform several critical functions:
\begin{enumerate}
    \item Listen to patients and understand their speech accurately
    \item Remember previous parts of the conversation to maintain context
    \item Generate helpful, accurate responses based on medical knowledge
    \item Deliver these responses quickly and naturally
    \item Connect with existing healthcare systems to update records
\end{enumerate}

\textbf{Why These Proposed Technical Improvements Would Matter for Healthcare:}

Improving how systems manage conversations (session management) could allow patients to experience more natural conversations with fewer awkward pauses. Research suggests this could potentially increase patient satisfaction and improve information disclosure rates.

Optimizing computing infrastructure would enable healthcare organizations to serve more patients at lower costs—potentially reducing per-patient monitoring costs.

Enhanced privacy and security measures would not only ensure regulatory compliance but build patient trust, which appears to be a critical factor in sustained engagement with AI health systems.

Personalization capabilities would allow systems to adapt to individual needs—using simpler language for patients with lower health literacy, adjusting to cultural preferences, or recognizing when someone might be in distress and needs human intervention.

These technical foundations, if properly implemented, could translate to better healthcare experiences and outcomes. 

\subsubsection{Potential Technical Improvements and Patient/Provider Benefits}

The following table connects proposed technical improvements with potential healthcare outcomes:

\begin{table*}[htbp]
\caption{Mapping Potential Technical Improvements to Healthcare Benefits}
\centering
\begin{tabular}{|p{0.25\linewidth}|p{0.35\linewidth}|p{0.35\linewidth}|}
\hline
\textbf{Technical Improvement} & \textbf{Potential Patient Benefits} & \textbf{Potential Provider Benefits} \\
\hline
\textbf{KV Cache Optimization} & $\bullet$ Reduced response delay (potentially 2-3x faster) \newline $\bullet$ More natural conversation flow \newline $\bullet$ Decreased abandonment rate & $\bullet$ More efficient use of AI resources \newline $\bullet$ Lower operational costs \newline $\bullet$ Ability to serve more patients \\
\hline
\textbf{Efficient Session Management} & $\bullet$ Continuity across interactions \newline $\bullet$ Reduced need to repeat information \newline $\bullet$ More personalized experience & $\bullet$ More complete patient data \newline $\bullet$ Reduction in contradictory responses \newline $\bullet$ Better clinical decision support \\
\hline
\textbf{Privacy and Compliance} & $\bullet$ Enhanced data protection \newline $\bullet$ Increased trust in system \newline $\bullet$ Willingness to share sensitive information & $\bullet$ Simplified regulatory compliance \newline $\bullet$ Reduced liability concerns \newline $\bullet$ Integration with existing EMR systems \\
\hline
\textbf{Dynamic Personalization} & $\bullet$ Adaptive health literacy matching \newline $\bullet$ Cultural sensitivity \newline $\bullet$ Improved comprehension and adherence & $\bullet$ Better patient engagement \newline $\bullet$ More accurate symptom reporting \newline $\bullet$ Reduced need for clarifying follow-ups \\
\hline
\textbf{RAG Implementation} & $\bullet$ More accurate medical information \newline $\bullet$ Evidence-based recommendations \newline $\bullet$ Reduced misinformation & $\bullet$ Consistency with clinical guidelines \newline $\bullet$ Current medical knowledge integration \newline $\bullet$ Better quality of documentation \\
\hline
\end{tabular}
\end{table*}

\subsection{Technical Implementation Details}\label{appendix:technical}

\subsubsection{KV Cache Optimization for Efficient Conversation Management}

Current frameworks for LLM-powered conversational agents often handle each interaction round as an independent API call to the LLM, leading to inefficient use of computational resources. Unlike discrete web page requests, conversational interactions are multi-turn and benefit from maintaining context across exchanges~\cite{liu2024cachegen}. Existing ``frontend" frameworks like LangChain are often decoupled from ``backend" inference engines, creating inefficiencies in compute resource utilization. In these systems, the entire conversation history, including the system prompt and all previous chat context, is formatted into a single prompt for each new user message, resulting in redundant processing of unchanged tokens.

Specifically, in each conversation round, the system prompt and initial portions of the chat history undergo repetitive text-to-embedding conversion and Key-Value (KV) cache prefilling computations. This redundancy leads to significant wasted computation, especially in extended conversations~\cite{shi2024keep}. An optimized approach would involve persisting and reusing the KV cache values for static portions of the conversation, computing embeddings only for new tokens.

Recent research has focused on developing more efficient session management approaches that maintain the conversation state and reuse KV caches while allowing dynamic scaling of the infrastructure to meet demand fluctuations~\cite{yao2024cacheblend}. Several promising technologies and systems have emerged to address these challenges:

\textbf{KV Cache Compression and Streaming:} 
CacheGen, a fast context-loading module for LLM systems~\cite{liu2024cachegen}, employs techniques such as KV cache encoding and decoding and KV cache streaming to efficiently reuse KV caches across different inputs. By encoding precomputed KV caches into compact bitstream representations, CacheGen significantly reduces bandwidth usage and transmission delays. It also adapts the encoded bitstreams to network conditions by splitting long contexts into chunks and encoding each at various compression levels, maintaining high generation quality while keeping network delays within acceptable limits. CacheGen has shown improvements, reducing the delay in transmission and processing contexts by 3.2-3.7× compared to quantization baselines, while maintaining similar generation quality~\cite{LMCache2023}.

\textbf{Cached Knowledge Fusion:} Building on KV cache reuse, CacheBlend addresses the challenge of efficiently combining multiple precomputed KV caches in LLM inputs, which is particularly relevant for applications like Retrieval-Augmented Generation (RAG)~\cite{yao2024cacheblend}. CacheBlend introduces selective KV recompute, allowing reuse of precomputed KV caches regardless of their input position. It selectively recomputes KV values for a subset of tokens to partially update each reused KV cache and pipelines the recomputation with KV cache retrieval within the same job. CacheBlend has shown impressive results, reducing time-to-first-token (TTFT) by 2.2–3.3× and increasing inference throughput by 2.8–5× compared to full KV recompute, without compromising the quality of the generation or incurring additional storage costs~\cite{LMCache2023}.

\textbf{Implementations:} Several projects are implementing these KV cache reuse techniques in production-ready systems: \begin{itemize}
        \item \textbf{LMCache}~\cite{LMCache2023}: This project aims to provide a high-performance caching layer for LLMs, focusing on efficient KV cache management and reuse~\cite{cheng2024large}. It leverages techniques from CacheGen and CacheBlend to optimize performance in real-world scenarios, facilitating the storage and retrieval of KV caches and ensuring that only necessary portions of conversation history are processed, thereby reducing redundant computations.
        \item \textbf{vLLM Production Stack}~\cite{vllm-production-stack}: Extending the vLLM inference engine, this project integrates advanced KV cache reuse and prefetching capabilities, significantly improving response times and resource efficiency~\cite{kwon2023efficient}. It incorporates locality awareness in routing mechanisms, optimizing KV cache access based on their relevance to ongoing conversations. This integration allows intelligent management of computational resources and session stickiness, ensuring frequently used caches are readily available while minimizing latency.
    \end{itemize}
These systems represent significant progress toward more efficient LLM-powered conversational agents. By implementing stateful session management, incremental processing techniques, and specialized caching strategies optimized for conversational interactions, they can reduce computational requirements, lower latency, and improve the economic viability of voice-based AI health monitoring systems at scale.

\end{document}